\PassOptionsToPackage{usenames,dvipsnames,table}{xcolor}
\documentclass[11pt,a4paper]{article}
\usepackage[draft]{hyperref}

\usepackage{naaclhlt2018}
\usepackage{times}

\usepackage[utf8]{inputenc} 
\usepackage[T1]{fontenc}    
\usepackage{url}            
\usepackage{booktabs}       
\usepackage{amsfonts}       
\usepackage{nicefrac}       
\usepackage{microtype}      
\usepackage{latexsym}
\usepackage{amssymb}
\usepackage{amsmath}
\usepackage{multirow}
\usepackage{rotating}
\usepackage{subfig}
\usepackage{array}
\usepackage{enumitem}
\usepackage{makecell}
\usepackage{xcolor}
\usepackage[toc,page]{appendix}
\usepackage{xspace}
\usepackage{mdwlist}
\usepackage{graphics}
\usepackage{color}
\usepackage{epsfig}
\usepackage{alltt}
\usepackage{moreverb}
\usepackage{fancyvrb}
\usepackage{enumerate}
\usepackage{colortbl}           
\usepackage{relsize}
\usepackage{txfonts}
\usepackage[labelfont=bf]{caption}
\usepackage{float}
\usepackage{etoolbox}
\usepackage{arydshln}
\usepackage{setspace}
\usepackage{mathtools}
\usepackage{bm}
\usepackage{soul}
\usepackage[draft]{pgf}
\usepackage[normalem]{ulem}
\usepackage{wasysym,textcomp}

\usepackage{tabularx}

\definecolor{vlgray}{gray}{0.95}

\newcolumntype{L}[1]{>{\raggedright\let\newline\\\arraybackslash\hspace{0pt}}m{#1}}
\newcolumntype{C}[1]{>{\centering\let\newline\\\arraybackslash\hspace{0pt}}m{#1}}
\newcolumntype{R}[1]{>{\raggedleft\let\newline\\\arraybackslash\hspace{0pt}}m{#1}}

\usepackage{lscape}
 \newcolumntype{b}{>{\hsize=2.3\hsize}X}
\newcolumntype{s}{>{\hsize=.45\hsize}X}
\newcolumntype{m}{>{\hsize=.9\hsize}X}

\newtoggle{draft}
\toggletrue{draft}
\iftoggle{draft}{
\newcommand{\roy}[1]{\textcolor{BurntOrange}{[#1 \textsc{--Roy}]}}
\newcommand{\wacomment}[1]{\textcolor{Blue}{[#1 \textsc{--Waleed}]}}
\newcommand{\bhavana}[1]{\textcolor{Green}{[#1 \textsc{--Bhavana}]}}
\newcommand{\dk}[1]{\textcolor{Maroon}{[#1 \textsc{--DK}]}}
}{
\newcommand{\roy}[1]{}
\newcommand{\wacomment}[1]{}
\newcommand{\bhavana}[1]{}
\newcommand{\dk}[1]{}
}

\newcommand{\camready}[1]{}
\aclfinalcopy

\newcommand{\com}[1]{}

\makeatletter
\g@addto@macro\normalsize{%
  \setlength\abovedisplayskip{1pt}
  \setlength\belowdisplayskip{1pt}
  \setlength\abovedisplayshortskip{1pt}
 \setlength\belowdisplayshortskip{1pt}
}
\renewcommand{\paragraph}{%
  \@startsection{paragraph}{4}%
  {\z@}{0.15ex \@plus 0.5ex \@minus .2ex}{-1em}%
  {\normalfont\normalsize\bfseries}%
}
\makeatother

\newcommand{\dataset}{\textsc{P}eer\textsc{R}ead\xspace}

\title{A Dataset of Peer Reviews (\dataset): \\
Collection, Insights and NLP Applications}

\author{
\makecell{Dongyeop Kang$^1$\quad Waleed Ammar$^2$\quad Bhavana Dalvi Mishra$^2$\quad Madeleine van Zuylen$^2$\quad \\Sebastian Kohlmeier$^2$\quad Eduard Hovy$^1$\quad Roy Schwartz$^{2,3}$} \\
$^1$School of Computer Science, Carnegie Mellon University, Pittsburgh, PA, USA \\
$^2$Allen Institute for Artificial Intelligence, Seattle, WA, USA\\
$^3$Paul G. Allen Computer Science \& Engineering, University of Washington, Seattle, WA, USA\\
{\tt $\{$dongyeok,hovy$\}$@cs.cmu.edu}\\
{\tt $\{$waleeda,bhavanad,madeleinev,sebastiank,roys$\}$@allenai.org}\\}

\date{}

\begin{document}
\maketitle

\begin{abstract}
Peer reviewing is a central component in the scientific publishing process.
We present the first public dataset of scientific peer reviews available for research purposes (\dataset v1),\footnote{\url{https://github.com/allenai/PeerRead}} providing an opportunity to study this important artifact.
The dataset consists of 14.7K paper drafts and the corresponding accept/reject decisions in top-tier venues including ACL, NIPS and ICLR. 
The dataset also includes 10.7K textual peer reviews written by experts for a subset of the papers.
We describe the data collection process and report  interesting observed phenomena in the peer reviews.
We also propose two novel NLP tasks based on this dataset and provide simple baseline models.
In the first task, we show that simple models can predict whether a paper is accepted with up to 21\% error reduction compared to the majority baseline.
In the second task, we predict the numerical scores of review aspects and show that simple models can outperform the mean baseline for aspects with high variance such as `originality' and `impact'.
\end{abstract}

\section{Introduction}

Prestigious scientific venues use peer reviewing to decide which papers to include in their journals or proceedings. 
While this process seems essential to scientific publication, it is often a subject of debate.
Recognizing the important consequences of peer reviewing, several researchers studied various aspects of the process, including consistency,  bias, author response and  general review quality  \cite[e.g.,][]{Greaves:2006,Ragone:2011,de2017preserving}.
For example, the organizers of the NIPS 2014 conference assigned 10\% of conference submissions to two different sets of reviewers to measure the consistency of the peer reviewing process, and observed that the two committees disagreed on the accept/reject decision for more than a quarter of the papers \cite{Langford:2015}. 

Despite these efforts, quantitative studies of peer reviews had been limited, for the most part, to the few individuals who had access to peer reviews of a given venue (e.g., journal editors and program chairs).
The goal of this paper is to lower the barrier to studying peer reviews for the scientific community by introducing the first public dataset of peer reviews for research purposes: \dataset. 

We use three strategies to construct the dataset:
(i) We collaborate with conference chairs and conference management systems to allow authors and reviewers to opt-in their paper drafts and peer reviews, respectively.
(ii) We crawl publicly available peer reviews and annotate textual reviews with numerical scores for aspects such as `clarity' and `impact'.
(iii) We crawl arXiv submissions which coincide with important conference submission dates and check whether a similar paper appears in proceedings of these conferences at a later date.
In total, the dataset consists of 14.7K paper drafts and the corresponding accept/reject decisions, including a subset of 3K papers for which we have 10.7K textual reviews written by experts.
We plan to make periodic releases of \dataset, adding more sections for new venues every year.
We provide more details on data collection in \S\ref{sec:data}.

The \dataset dataset can be used in a variety of ways.
A quantitative analysis of the peer reviews can provide insights to help better understand (and potentially improve) various nuances of the review process.
For example, in \S\ref{sec:insights}, we analyze correlations between the overall recommendation score and individual aspect scores (e.g., clarity, impact and originality) and quantify how reviews recommending an oral presentation differ from those recommending a poster.
Other examples might include aligning review scores with authors to reveal gender or nationality biases.
From a pedagogical perspective, the \dataset dataset also provides inexperienced authors and first-time reviewers with diverse examples of peer reviews.

As an NLP resource, peer reviews raise interesting challenges, both from the realm of sentiment analysis---predicting various properties of the reviewed paper, e.g., clarity and novelty, as well as that of text generation---given a paper, automatically generate its review.
Such NLP tasks, when solved with sufficiently high quality, might help reviewers, area chairs and program chairs in the reviewing process, e.g., by lowering the number of reviewers needed for some paper submission.

In \S\ref{sec:applications}, we introduce two new NLP tasks based on this dataset:
(i) predicting whether a given paper would be accepted to some venue, and
(ii) predicting the numerical score of certain aspects of a paper.
Our results show that we can predict the accept/reject decisions with 6--21\% error reduction compared to the majority reject-all baseline, in four different sections of \dataset. 
Since the baseline models we use are fairly simple, there is plenty of room to develop stronger models to make better predictions. 



\section{Peer-Review  Dataset (\dataset)}\label{sec:data}




Here we describe the collection and compilation of \dataset, our scientific peer-review dataset.
For an overview of the dataset, see Table \ref{tab:dataset}. 

\begin{table}[t]
\tabcolsep=0.11cm
\centering\vspace{-1mm}
\begin{small}
\begin{tabular}{@{}r|rrcr@{}}
\toprule
\textbf{Section}  & \textbf{\#Papers} & \textbf{\#Reviews} & \textbf{Asp.} &  {\textbf{Acc / Rej}} \\
\midrule
NIPS 2013--2017	&2,420 & 9,152 		&\texttimes				& 2,420 / 0 \\
ICLR 2017	&427  & 1,304 		&\checked\com{$^{*}$}				& 172 / 255 \\ \hline
ACL 2017	&137 & 275 			&\checked				& 88 / 49 \\
CoNLL 2016	&22 &39 			&\checked				& 11 / 11 \\ \hline 
arXiv 2007--2017	&11,778 	& --- 		& ---			&2,891 / 8,887 \\ 
\hline \hline
\textit{total}	&14,784 & 10,770	&			& \\ 
\bottomrule
\end{tabular}
\end{small}
\caption{\label{tab:dataset} The \dataset dataset. 
\textbf{Asp.}~indicates whether the reviews have aspect specific scores (e.g., clarity). 
Note that ICLR contains the aspect scores assigned by our annotators (see Section \ref{sec:annotation}).
\textbf{Acc/Rej} is the distribution of accepted/rejected papers.
Note that NIPS provide reviews only for accepted papers.
}
\end{table}


\subsection{Review Collection}
Reviews in \dataset belong to one of the two categories:

\paragraph{Opted-in reviews.}
We coordinated with the Softconf conference management system and the conference chairs for CoNLL 2016\footnote{The 20$^{\text{th}}$ SIGNLL Conference on Computational Natural Language Learning; \url{http://www.conll.org/2016}}
and ACL 2017\footnote{The 55$^{\text{th}}$ Annual Meeting of the Association for Computational Linguistics; \url{http://acl2017.org/}}
conferences to allow authors and reviewers to opt-in their drafts and reviews, respectively, 
 to be included in this dataset.
A submission is included only if (i) the corresponding author opts-in the paper draft, and (ii) at least one of the reviewers opts-in their anonymous reviews.
This resulted in 39 reviews for 22 CoNLL 2016 submissions, and 275 reviews for 137 ACL 2017 submissions. 
Reviews include both text and aspect scores (e.g., calrity) on a scale of 1--5. 

\paragraph{Peer reviews on the web.}
In 2013, the NIPS conference\footnote{The Conference on Neural Information Processing Systems; \url{https://nips.cc/}} began attaching all accepted papers with their anonymous textual review comments, as well as a confidence level on a scale of 1--3.
We collected all accepted papers and their reviews for NIPS 2013--2017, a total of 9,152 reviews for 2,420 papers.

Another source of reviews is the OpenReview  platform:\footnote{\url{http://openreview.net}} a conference management system which promotes open access and open peer reviewing.
Reviews include text, as well as numerical recommendations between 1--10 and confidence level between 1--5. 
We collected all submissions to the ICLR 2017 conference,\footnote{The 5$^{\text{th}}$ International Conference on Learning Representations; \url{https://iclr.cc/archive/www/2017.html}} a total of 1,304 official, anonymous reviews 
for 427 papers (177 accepted and 255 rejected).\footnote{The platform also allows any person to review the paper by adding a comment, but we only use the official reviews of reviewers assigned to review that paper.}

\subsection{arXiv Submissions}\label{sec:arxiv}
arXiv\footnote{\url{https://arxiv.org/}} is a popular platform for pre-publishing research in various scientific fields including physics, computer science and biology.
While arXiv does not contain reviews, we automatically label a subset of arXiv submissions in the years 2007--2017 (inclusive)\footnote{For consistency, we only include the first arXiv version of each paper (accepted or rejected) in the dataset.} as accepted or probably-rejected, with respect to a group of top-tier NLP, ML and AI venues: ACL, EMNLP, NAACL, EACL, TACL, NIPS, ICML, ICLR and AAAI. 

\paragraph{Accepted papers.}
In order to assign `accepted' labels, we use the dataset provided by \citet{Sutton:2017} who matched arXiv submissions to their bibliographic entries in the DBLP directory\footnote{\url{http://dblp.uni-trier.de/}} by comparing titles and author names using Jaccard's distance. 
To improve our coverage, we also add an arXiv submission if its title matches an accepted paper in one of our target venues with a relative Levenshtein distance \cite{Levenshtein:1966} of < 0.1.
This results in a total of 2,891 accepted papers.

\paragraph{Probably-rejected papers.}
We use the following criteria to assign a `probably-rejected' label for an arXiv submission: 
\begin{itemize}[noitemsep,topsep=0pt,leftmargin=*]
\item The paper wasn't accepted to any of the target venues.\footnote{Note that some of the `probably-rejected' papers may be published at workshops or other venues.}
\item The paper was submitted to one of the arXiv categories \texttt{cs.cl}, \texttt{cs.lg} or \texttt{cs.ai}.\footnote{See \url{https://arxiv.org/archive/cs} for a description of the computer science categories in arXiv.}
\item The paper wasn't cross-listed in any non-\texttt{cs} categories.
\item The submission date\footnote{If a paper has multiple versions, we consider the submission date of the first version.} was within one month of the submission deadlines of our target venues (before or after).
\item The submission date coincides with at least one of the arXiv papers accepted for one of the target venues. 
\end{itemize}

This process results in 8,887 `probably-rejected' papers.



\paragraph{Data quality.}
We did a simple sanity check in order to estimate the number of papers that we labeled as `probably-rejected', but were in fact accepted to one of the target venues.
Some authors add comments to their arXiv submissions to indicate the publication venue.
We identified arXiv papers with a comment which matches the term ``accept'' along with any of our target venues (e.g., ``nips''), but not the term ``workshop''. 
We found 364 papers which matched these criteria, 
352 out of which were labeled as `accepted'.
Manual inspection of the remaining 12 papers showed that one of the papers was indeed a false negative (i.e., labeled as `probably-rejected' but accepted to one of the target venues) due to a significant change in the paper title.
The remaining 11 papers were not accepted to any of the target venues (e.g., ``accepted at WMT@ACL 2014'').

\subsection{Organization and Preprocessing}
We organize v1.0 of the \dataset dataset in five sections: CoNLL 2016, ACL 2017, ICLR 2017, NIPS 2013--2017 and arXiv 2007--2017.\footnote{We plan to periodicly release new versions of \dataset.}
Since the data collection varies across sections, different sections may have different license agreements.
The papers in each section are further split into standard training, development and test sets with 0.9:0.05:0.05 ratios.
In addition to the PDF file of each paper, we also extract its textual content using the \texttt{Science Parse} library.\footnote{\url{https://github.com/allenai/science-parse}}
We represent each of the splits as a json-encoded text file with a list of paper objects, each of which consists of paper details, accept/reject/probably-reject decision, and a list of reviews.

\subsection{Aspect Score Annotations}\label{sec:annotation}
In many publication venues, reviewers assign numeric aspect scores (e.g., clarity, originality, substance) as part of the peer review.
Aspect scores could be viewed as a structured summary of the strengths and weaknesses of a paper.
While aspect scores assigned by reviewers are included in the opted-in sections in \dataset, they are missing from the remaining reviews.
In order to increase the utility of the dataset, we annotated 1.3K reviews with aspect scores, based on the corresponding review text.
Annotations were done by two of the authors. 
In this subsection, we describe the annotation process in detail.

\paragraph{Feasibility study.} 
As a first step, we verified the feasibility of the annotation task by annotating nine reviews for which aspect scores are available.
The annotators were able to infer about half of the aspect scores from the corresponding review text (the other half was not discussed in the review text). 
This is expected since reviewer comments often focus on the key strengths or weaknesses of the paper and are not meant to be a comprehensive assessment of each aspect.
On average, the absolute difference between our annotated scores and the gold scores originally provided by reviewers is 0.51 (on a 1--5 scale, considering only those cases where the aspect was discussed in the review text). 

\paragraph{Data preprocessing.} 
We used the official reviews in the ICLR 2017 section of the dataset for this annotation task.
We excluded unofficial comments contributed by arbitrary members of the community, comments made by the authors in response to other comments, as well as ``meta-reviews'' which state the final decision on a paper submission.
The remaining 1,304 official reviews are 
all written by anonymous reviewers assigned by the program committee to review a particular submission. 
We randomly reordered the reviews before annotation so that the annotator judgments based on one review are less affected by other reviews of the same paper.

\paragraph{Annotation guidelines.}
We annotated seven aspects for each review: appropriateness, clarity, originality, soundness/correctness, meaningful comparison, substance, and impact.
For each aspect, we provided our annotators with the instructions given to ACL 2016 reviewers for this aspect.\footnote{Instructions are provided in Appendix \ref{ssec:instructions}.} 
Our annotators' task was to read the detailed review text (346 words on average) and select a score between 1--5 (inclusive, integers only) for each aspect.\footnote{Importantly, our annotators only considered the review text, and did not have access to the papers.}
When review comments do not address a specific aspect, we do not select any score for that aspect, and instead use a special ``not discussed'' value.\com{\footnote{As discussed earlier, our feasibility study showed that review comments only discuss a subset of the aspects. Requiring the annotators to select a score would result in many low-confidence, unuseful annotations. Therefore, we decided it was necessary to add a ``not discussed'' option.}}

\paragraph{Data quality.}
In order to assess annotation consistency, the same annotators re-annotated a random sample consisting of 30 reviews.
On average, 77\% of the annotations were consistent (i.e., the re-annotation was exactly the same as the original annotation, or was off by 1 point) and 2\% were inconsistent (i.e., the re-annotation was off by 2 points or more).
In the remaining 21\%, the aspect was marked as ``not discussed'' in one annotation but not in the other.
We note that different aspects are discussed in the textual reviews at different rates.
For example, about 49\% of the reviews discussed the `originality' aspect, while only 5\% discussed `appropriateness'.

\section{Data-Driven Analysis of Peer Reviews}\label{sec:insights} \label{sec:analysis}
In this section, we showcase the potential of using \dataset for data-driven analysis of peer reviews.

\paragraph{Overall recommendation vs.~aspect scores.}
A critical part of each review is the overall recommendation score, a numeric value which best characterizes a reviewer's judgment of whether the draft should be accepted for publication in this venue.
While aspect scores (e.g., clarity, novelty, impact) help explain a reviewer's assessment of the submission, it is not necessarily clear which aspects reviewers appreciate the most about a submission when considering their overall recommendation.

To address this question, we measure pair-wise correlations between the overall recommendation and various aspect scores in the ACL 2017 section of \dataset and report the results in Table \ref{tab:overall_vs_aspects}.

\begin{table}[h]
\centering
\begin{center}
\begin{tabular}{r|c} 
\toprule
\textbf{Aspect} & \bm{$\rho$} \\ 
\midrule
Substance & 0.59 \\ 
Clarity & 0.42 \\ 
Appropriateness & 0.30 \\ 
Impact & 0.16 \\ 
Meaningful comparison & 0.15 \\ 
Originality & 0.08 \\ 
Soundness$/$Correctness & 0.01 \\ 
\bottomrule
\end{tabular}
\end{center}
\caption{\label{tab:overall_vs_aspects} Pearson's correlation coefficient $\rho$ between the overall recommendation and various aspect scores in the ACL 2017 section of \dataset. }
\end{table}

The aspects which correlate most strongly with the final recommendation are substance (which concerns the amount of work rather than its quality) and clarity.
In contrast, soundness/correctness and originality are least correlated with the final recommendation.
These observations raise interesting questions about what we collectively care about the most as a research community when evaluating paper submissions.

\paragraph{Oral vs.~poster.}
In most NLP conferences, accepted submissions may be selected for an oral presentation or a poster presentation.
The presentation format decision of accepted papers is based on recommendation by the reviewers.
In the official blog of ACL 2017,\footnote{\url{https://acl2017.wordpress.com/2017/03/23/conversing-or-presenting-poster-or-oral/}} the program chairs recommend that reviewers and area chairs make this decision based on the expected size of interested audience and whether the ideas can be grasped without back-and-forth discussion. 
However, it remains unclear what criteria are used by reviewers to make this decision.

To address this question, we compute the mean aspect score in reviews which recommend an oral vs.~poster presentation in the ACL 2017 section of \dataset, and report the results in Table \ref{tab:oral_vs_poster}.
Notably, the average `overall recommendation' score in reviews recommending an oral presentation is 0.9 higher than in reviews recommending a poster presentation, suggesting that reviewers tend to recommend oral presentation for submissions which are holistically stronger.

\begin{table}[h]
\centering
\begin{center}
\tabcolsep=0.05cm
\begin{tabular}{@{}r||cc||cc@{}}
\toprule
\textbf{Presentation format} & \textbf{Oral} & \textbf{Poster} & $\Delta$ & \textbf{stdev} \\
\midrule
Recommendation       & 3.83 & 2.92   & 0.90   & 0.89 \\
Substance            & 3.91 & 3.29   & 0.62   & 0.84 \\ 
Clarity              & 4.19 & 3.72   & 0.47   & 0.90 \\ 
Meaningful comparison& 3.60 & 3.36   & 0.24   & 0.82 \\ 
Impact               & 3.27 & 3.09   & 0.18   & 0.54 \\ 
Originality          & 3.91 & 3.88   & 0.02   & 0.87 \\ 
Soundness/Correctness&3.93 & 4.18   & -0.25  & 0.91 \\ 
\bottomrule
\end{tabular}
\end{center}
\caption{\label{tab:oral_vs_poster} Mean review scores for each presentation format (oral vs.~poster). Raw scores range between 1--5. For reference, the last column shows the sample standard deviation based on all reviews. }
\end{table}

\paragraph{ACL 2017 vs.~ICLR 2017.}
Table \ref{tab:acl2017_vs_iclr2017} reports the sample mean and standard deviation of various measurements based on reviews in the ACL 2017 and the ICLR 2017 sections of \dataset.
Most of the mean scores are similar in both sections, with a few notable exceptions.
The comments in ACL 2017 reviews tend to be about 50\% longer than those in the ICLR 2017 reviews.
Since review length is often thought of as a measure of its quality, this raises interesting questions about the quality of reviews in ICLR vs.~ACL conferences.
We note, however, that ACL 2017 reviews were explicitly opted-in while the ICLR 2017 reviews include all official reviews, which is likely to result in a positive bias in review quality of the ACL reviews included in this study.

Another interesting observation is that the mean appropriateness score is lower in ICLR 2017 compared to ACL 2017.
While this might indicate that ICLR 2017 attracted more irrelevant submissions, 
this is probably an artifact of our annotation process: reviewers probably only address appropriateness explicitly in their review if the paper is inappropriate,
which leads to a strong negative bias against this category in our ICLR dataset.

\begin{table}[h]
\centering
\begin{center}
\begin{tabular}{@{}r||c|c@{}}
\toprule
\textbf{Measurement} & \textbf{ACL'17} & \textbf{ICLR'17} \\
\midrule
Review length (words)   & $531{\pm 323}$ & $346{\pm 213}$ \\ 
Appropriateness         & $4.9{\pm 0.4}$ & $2.6{\pm 1.3}$ \\ 
Meaningful comparison   & $3.5{\pm 0.8}$ & $2.9{\pm 1.1}$ \\ 
Substance               & $3.6{\pm 0.8}$ & $3.0{\pm 0.9}$ \\ 
Originality             & $3.9{\pm 0.9}$ & $3.3{\pm 1.1}$ \\ 
Clarity                 & $3.9{\pm 0.9}$ & $4.2{\pm 1.0}$ \\ 
Impact                  & $3.2{\pm 0.5}$ & $3.4{\pm 1.0}$ \\ 
Overall recommendation  & $3.3{\pm 0.9}$ & $3.3{\pm 1.4}$ \\
\bottomrule
\end{tabular}
\end{center}
\caption{\label{tab:acl2017_vs_iclr2017} Mean $\pm$ standard deviation of various measurements on reviews in the ACL 2017 and ICLR 2017 sections of \dataset. 
Note that ACL aspects were written by the reviewers themselves, while ICLR aspects were predicted by our annotators based on the review.}
\end{table}

\com{We emphasize that the peer reviews analysis discussed earlier is subject to limitations of the dataset collection. 
For example, the ACL'17 opted-in reviews may exhibit different characteristics than other reviews in the same conference.
For more details on data collection, see \S\ref{sec:data}.}

\section{NLP Tasks}\label{sec:applications}\label{sec:experiments}

Aside from quantitatively analyzing peer reviews, \dataset can also be used to define interesting NLP tasks.
In this section, we introduce two novel tasks based on the \dataset dataset.
In the first task, given a paper draft, we predict whether the paper will be accepted to a set of target conferences.
In the second task, given a textual review, we predict the aspect scores for the paper such as novelty, substance and meaningful comparison.\footnote{We also experiment with conditioning on the paper itself to make this prediction.} 

Both these tasks are not only challenging from an NLP perspective, but also have potential applications. 
For example, models for predicting the accept/reject decisions of a paper draft might be used in recommendation systems for arXiv submissions.
Also, a model trained to predict the aspect scores given review comments using thousands of training examples might result in better-calibrated scores.

\subsection{Paper Acceptance Classification}

Paper acceptance classification is a binary classification task: given a paper draft, predict whether the paper will be accepted or rejected for a predefined set of venues.

\paragraph{Models.}
We train a binary classifier to estimate the probability of accept vs.~reject given a paper, i.e., $P (\texttt{accept=True} \mid \texttt{paper})$.
We experiment with different types of classifiers: logistic regression, SVM with linear or RBF kernels, Random Forest, Nearest Neighbors, Decision Tree, Multi-layer Perceptron, AdaBoost, and Naive Bayes.
We use hand-engineered features, instead of neural models, because they are easier to interpret.

We use 22 coarse features, e.g., length of the title and whether jargon terms such as `deep' and `neural' appear in the abstract, as well as sparse and dense lexical features.
The full feature set is detailed in Appendix \ref{ssec:features}.

\paragraph{Experimental setup.}
We experiment with the ICLR 2017 and the arXiv sections of the \dataset dataset. We train separate models for each of the arXiv category: \texttt{cs.cl}, \texttt{cs.lg}, and \texttt{cs.ai}.
We use python's sklearn's implementation of all models \cite{scikit-learn}.\footnote{\url{http://scikit-learn.org/stable/}}
We consider various regularization parameters for SVM and logistic regression (see Appendix \ref{ssec:hyperparameters} for a detailed description of all hyperparameters).
We use the standard test split and tune our hyperparameters using 5-fold cross validation on the training set.

\paragraph{Results.}
\begin{table}[t]
\tabcolsep=0.11cm
\centering
\begin{tabular}{@{}l@{}|c|ccc@{}}
\toprule
        & \textbf{ICLR} & \com{\textbf{arXiv} &} \textbf{\texttt{cs.cl}} & \textbf{\texttt{cs.lg}} & \textbf{\texttt{cs.ai}} \\
\midrule
Majority & 57.6 & \com{77.2 &} 68.9 & 67.9 & 92.1 \\ \hline
\makecell{Ours \\($\Delta$)} & \makecell{65.3 \\+7.7} & \com{\makecell{79.1 \\+1.8} &} \makecell{75.7 \\+6.8} & \makecell{70.7 \\+2.8} & \makecell{92.6 \\+0.5}\\
\bottomrule
\end{tabular}
\caption{\label{tab:acceptance} Test accuracies (\%) for acceptance classification. Our best model outperforms the majority classifiers in all cases.}
\end{table}

Table~\ref{tab:acceptance} shows our test accuracies for the paper acceptance task.
Our best model outperforms the majority classifier in all cases, 
with up to 22\% error reduction.
Since our models lack the sophistication to assess the quality of the work discussed in the given paper, this might indicate that some of the features we define are correlated with strong papers, or bias reviewers' judgments.

\com{
\begin{table}[t]
\centering
\begin{tabular}{@{}l|cc@{}}
\toprule
        & \textbf{ICLR} & \textbf{ArXiv}  \\
\midrule
Majority &  57.6 & 77.2\\
\hline
Lexical (w2v+tfidf) + coarse 					&  61.5 & 78.2\\
Lexical (w2v) + coarse						&  \textbf{65.3} & 78.6\\
Lexical (bow+tfidf) + coarse					&  56.4 & \textbf{79.1}\\
\hline
Coarse only     					& 61.5 & 78.2\\
Lexical only (w2v+tfidf)  			& 57.6 & 77.3 \\
Lexical only (bow+tfidf)  			& 58.9 & 78.0\\
\bottomrule
\end{tabular}
\caption{\label{tab:ablation} Ablation study for paper acceptance prediction.
The table shows the performance of our model with different groups of features. }
\end{table}}

We run an ablation study for this task for the ICLR and arXiv sections\com{, showing the best performing model}.
We train only one model for all three categories in arXiv to simplify our analysis.
Table~\ref{tab:top-features} shows the absolute degradation in test accuracy of the best performing model when we remove one of the features. 
The table shows that some features have a large contribution on the classification decision:
adding an appendix, a large number of theorems or equations, 
the average length of the text preceding a citation, 
the number of papers cited by this paper that were published in the five years before the submission of this paper,
whether the abstract contains a phrase ``state of the art'' for ICLR or ``neural'' for arXiv, 
and length of title.\footnote{Coefficient values of each feature are provided in Appendix \ref{ssec:features}.}

\begin{table}[t]
\small
\centering
\tabcolsep=0.11cm

\parbox{.43\linewidth}{
\centering
\begin{tabular}{@{}lr@{}}\hline
\toprule
\textbf{ICLR} & \textbf{\%} \\
\midrule
Best model & 65.3\\
-- appendix & --5.4\\
-- num\_theorems	 & 	--3.8\\
-- num\_equations	 & 	--3.8\\
-- avg\_len\_ref	 &  --3.8\\
-- abstract$_{\text{state-of-the-art}}$	 & 	--3.5\\
-- \#recent\_refs	 & 	 --2.5\\
\bottomrule
\end{tabular}
}
\quad\quad\quad
\parbox{.43\linewidth}{
\begin{tabular}{@{}lr@{}}\hline
\toprule
\textbf{arXiv} & \textbf{\%} \\
\midrule
Best model & 79.1\\
-- avg\_len\_ref	 & 	--1.4 \\
-- num\_uniq\_words	 & 	--1.1 \\
-- num\_theorems	 & 	--1.0\\
-- abstract$_{\text{neural}}$	 & 	--1.0\\
-- num\_refmentions	 & 	--1.0\\
-- title\_length & --1.0\\
\bottomrule
\end{tabular}
}
\caption{\label{tab:top-features} 
The absolute \% difference in accuracy on the paper acceptance prediction task when we remove only one feature from the full model.
Features with larger negative differences are more salient, and we only show the six most salient features for each section.
The features are
num\_${X}$: number of $X$ (e.g., theorems or equations),
avg\_len\_ref: average length of context before a reference,
appendix: does paper have an appendix,
abstract$_{X}$: does the abstract contain the phrase $X$, 
num\_uniq\_words: number of unique words,
num\_refmentions: number of reference mentions, and
\#recent\_refs: number of cited papers published in the last five years.
}
\end{table}



\com{
\begin{figure}[t]	
\centering
{
\includegraphics[trim=2cm 0.6cm 0cm 1cm,clip,width=.99\linewidth]{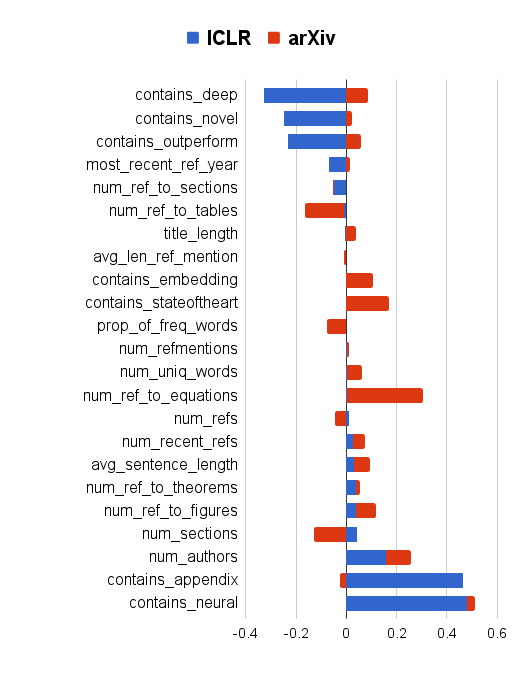}
}
\caption{\label{fig:coef} Coefficient on coarse features: ICLR (blue) and arXiv (red)}
\end{figure}
}

\subsection{Review Aspect Score Prediction}
\begin{figure*}[t]	
\centering
{
\includegraphics[trim=0cm -0.7cm 0cm 0cm,clip,width=.49\linewidth]{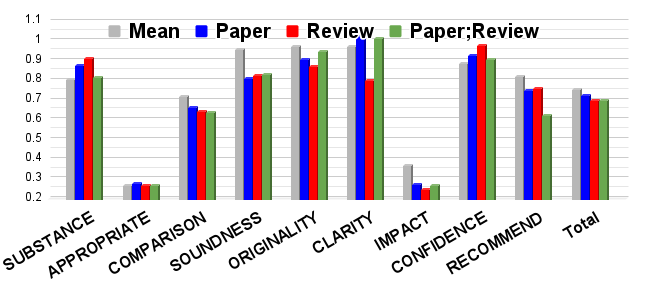}
\includegraphics[trim=0cm 0cm 0cm 0cm,clip,width=.49\linewidth]{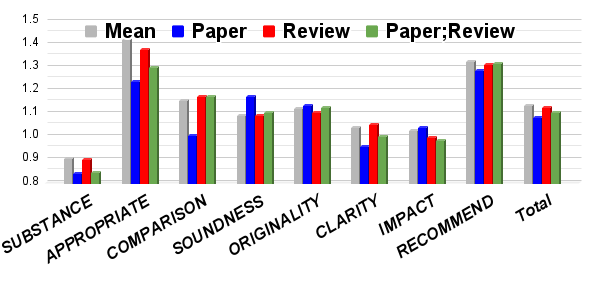}
}
\caption{\label{fig:rmse} Root mean squared error (RMSE, lower is better) on the test set for the aspect prediction task on the ACL 2017 (left) and the ICLR 2017 (right) sections of \dataset.}
\end{figure*}

The second task is a multi-class regression task to predict scores for seven review aspects:
`impact', `substance', `appropriateness', `comparison', `soundness', `originality' and `clarity'. 
For this task, we use the two sections of \dataset which include aspect scores: ACL 2017 and ICLR 2017.\footnote{The CoNLL 2016 section also includes aspect scores but is too small for training.}


\paragraph{Models.}
We use a regression model which predicts a floating-point score for each aspect of interest given a sequence of tokens.
We train three variants of the model to condition on (i) the paper text only, (ii) the review text only, or (iii) both paper and review text.

We use three neural architectures:
convolutional neural networks \cite[CNN,][]{Zhang:2015}, recurrent neural networks \cite[LSTM,][]{Hochreiter:1997}, and deep averaging networks \cite[DAN,][]{Iyyer:2015}.
In all three architectures, we use a linear output layer to make the final prediction.
The loss function is the mean squared error between predicted and gold scores.
We compare against a baseline which always predicts the mean score of an aspect, computed on the training set.\footnote{This baseline is guaranteed to obtain mean square errors less than or equal to the majority baseline.}

\paragraph{Experimental setup.}
We train all models on the standard training set for 100 iterations, and select the best performing model on the standard development set.
We use a single 100 dimension layer LSTM and CNN, and a single output layer of 100 dimensions for all models.
We use GloVe 840B embeddings \cite{pennington2014glove} as input word representations, without tuning, and keep the 35K most frequent words and replace the rest with an UNK vector.
The CNN model uses 128 filters and 5 kernels.
We use an RMSProp optimizer \cite{Tieleman:2012} with 0.001 learning rate, 0.9 decay rate, 5.0 gradient clipping, and a batch size of 32.
Since scientific papers tend to be long, we only take the first 1000 and 200 tokens of each paper and review, respectively, and concatenate the two prefixes when the model conditions on both the paper and review text.\footnote{We note that the goal of this paper is to demonstrate potential uses of \dataset, rather than develop the best model to address this task, which explains the simplicity of the models we use.}

\paragraph{Results.}
Figure~\ref{fig:rmse} shows the test set root mean square error (RMSE) on the aspect prediction task (lower is better).
For each section (ACL 2017 and ICLR 2017), and for each aspect, we report the results of four systems: `Mean' (baseline), `Paper', `Review' and `Paper;Review' (i.e., which information the model conditions on). 
For each variant, the model which performs best on the development set is selected.

We note that aspects with higher RMSE scores for the `Mean' baseline indicate higher variance among the review scores for this aspect, so we focus our discussion on these aspects. 
In the ACL 2017 section, the two aspects with the highest variance are `originality' and `clarity'.
In the ICLR 2017 section, the two aspects with the highest variance are `appropriateness' and `meaningful comparison'.
Surprisingly, the `Paper;Review' model outperforms the `Mean' baseline in all four aspects, and the `Review' model outperforms the `Mean' baseline in three out of four.
On average, all models slightly improve over the `Mean' baseline.




\com{
\begin{table}[h]
\centering\small
\begin{tabular}{@{}r|ccc|ccc@{}}
\toprule
&  \multicolumn{3}{c|}{ACL} & \multicolumn{3}{c}{ICLR} \\
\midrule
mean & \multicolumn{3}{c|}{0.749} &  \multicolumn{3}{c}{1.133}\\
Major &\multicolumn{3}{c|}{0.848}& \multicolumn{3}{c}{1.409}\\
\hline
& \texttt{p} & \texttt{r} & \texttt{p;r} & \texttt{p} & \texttt{r} & \texttt{p;r}\\
\midrule
\textit{CNN} & 0.795 & 0.751 & 0.760 & \textbf{1.141} & 1.165 & 1.146\\
\textit{LSTM} & 0.781 & 0.752 & \textbf{0.725} & 1.150 & 1.152 & 1.148\\
\textit{DAN} & 0.776 & 0.850 & 0.762 & 1.144 & 1.126 & 1.210 \\
\bottomrule
\end{tabular}
\caption{\label{tab:predict2} Total averaged root mean squared error (RMSE) test score across all aspects between different encoders (\textit{CNN}, \textit{RNN}, \textit{DAN}) and between encoding features (\texttt{p} for paper, \texttt{r} for review, \texttt{p;r} for both).}

\end{table}

Finally, Table~\ref{tab:predict2} compares between the different models and between different text types.
If only small length of text is given like \texttt{review}, simple model such as \textit{DAN} is likely underfitting to the data. 
If both \texttt{review} and \texttt{paper} are given, \textit{DAN} outperforms the other complicated models.
}
\com{
\dk{Comparison table between training separate model for each aspect and training combined model for predicting entire aspects}\wacomment{I think adding such comparison may confuse readers. Why is it important to show this comparison?}
}
\section{Related Work}\label{sec:related}
Several efforts have recently been made to collect peer reviews. 
Publons\footnote{\url{publons.com/dashboard/records/review/}} consolidates peer reviews data to build public reviewer profiles for participating reviewers. 
Crossref maintains the database of DOIs for its 4000+ publisher members. 
They recently launched a service to add peer reviews as part of metadata for the scientific articles.\footnote{\url{https://www.crossref.org/blog/peer-reviews-are-open-for-registering-at-crossref/}}
Surprisingly, however, most of the reviews are not made publicly available.
In contrast, we collected and organized \dataset such that it is easy for other researchers to use it for research purposes, replicate experiments and make a fair comparison to previous results.


There have been several efforts to analyze the peer review process \cite[e.g.,][]{Bonaccorsi2015JournalRA,rennie2016make}. 
Editors of the British Journal of Psychiatry found differences in courtesy between signed and unsigned reviews \cite{walsh:2000}.
\newcite{Ragone:2011} and \newcite{Birukou2011AlternativesTP} analyzed ten CS conferences and found low correlation between review scores and the impact of papers in terms of future number of citations.
\newcite{fang2016nih} presented similar observations for NIH grant application reviews and their productivity.
\citet{Langford:2015} pointed to inconsistencies in the peer review process. 

Several recent venues had single vs.~double blind review experiments, which pointed to single-blind reviews leading to increased biases towards male authors \cite{Roberts:2016} and famous institutions \cite{Tomkins:2017}.
Further, \citet{LeGoues:2017} showed that reviewers are unable to successfully guess the identity of the author in a double-blind review.
Recently, there have been several initiatives by program chairs in major NLP conferences to study various aspects of the review process, mostly author response and general review quality.\footnote{See \url{https://nlpers.blogspot.com/2015/06/some-naacl-2013-statistics-on-author.html} and \url{https://acl2017.wordpress.com/2017/03/27/author-response-does-it-help/}}
In this work, we provide a large scale dataset that would enable the wider scientific community to further study the properties of peer review, and potentially come up with enhancements to current peer review model.

Finally, the peer review process is meant to judge the quality of research work being disseminated to the larger research community. 
With the ever-growing rates of articles being submitted to top-tier conferences in Computer Science and pre-print repositories \cite{Sutton:2017}, there is a need to expedite the peer review process. 
\citet{Balachandran:2013} proposed a method for automatic analysis of conference submissions to recommend relevant reviewers.
Also related to our acceptance predicting task are \cite{Tsur:2009} and \newcite{Ashok:2013}, both of which focuses on predicting book reviews.
Various automatic tools like Grammerly\footnote{\url{https://www.grammarly.com/}} can assist reviewers in discovering grammar and spelling errors. 
Tools like Citeomatic\footnote{\url{http://allenai.org/semantic-scholar/citeomatic/}} \cite{chandra2017Citeomatic} are especially useful in finding relevant articles not cited in the manuscript.
We believe that the NLP tasks presented in this paper, predicting the acceptance of a paper and the aspect scores of a review, 
can potentially serve as useful tools for writing a paper, reviewing it, and deciding about its acceptance.

\section{Conclusion}\label{sec:conclusion}
We introduced \dataset, the first publicly available peer review dataset for research purposes, containing 14.7K papers and 10.7K reviews.
We analyzed the dataset, showing interesting trends such as a high correlation between overall recommendation and recommending an oral presentation.
We defined two novel tasks based on \dataset: (i) predicting the acceptance of a paper based on textual features and 
(ii) predicting the score of each aspect in a review based on the paper and review contents. 
Our experiments show that certain properties of a paper, such as having an appendix, are correlated with higher acceptance rate.
Our primary goal is to motivate other researchers to explore these tasks and develop better models that outperform the ones used in this work.
More importantly, we hope that other researchers will identify novel opportunities which we have not explored to analyze the peer reviews in this dataset. 
As a concrete example, it would be interesting to study if the accept/reject decisions reflect author demographic biases (e.g., nationality).

\com{
Our work has several future works in many directions:
\begin{itemize}[noitemsep,topsep=0pt,leftmargin=*]
\item \textit{Applications}: we haven't explored much on our review text yet. Studying on generating sentiment reviews or aspect specific reviews given a paper would be bit ambitious but interesting applications.
\item \textit{Analysis}: Our analysis on acceptance shows some interesting features for paper acceptance. With our science-parse data, someone can find another interesting features for paper acceptance. For example, total number of citation on the referenced papers, grammar checking, discourse related features, and so on.
\item \textit{Coverage}: Collecting both positive and negative reviews are indeed expensive and difficult. Based on our current dataset, we will extend coverage of our review collection by asking conference chairs for sharing their reviews or collecting more open reviews in public. 
\item \textit{Service}: Our analysis could be integrated to existing academic paper search engines for additional features for both authors and reviewers. 
\end{itemize}}

\section*{Acknowledgements}
This work would not have been possible without the efforts of Rich Gerber and Paolo Gai (developers of the \url{softconf.com} conference management system), Stefan Riezler, Yoav Goldberg (chairs of CoNLL 2016), Min-Yen Kan, Regina Barzilay (chairs of ACL 2017) for allowing authors and reviewers to opt-in for this dataset during the official review process.
We thank the \url{openreview.net}, \url{arxiv.org} and \url{semanticscholar.org} teams for their commitment to promoting transparency and openness in scientific communication. 
We also thank Peter Clark, Chris Dyer, Oren Etzioni, Matt Gardner, Nicholas FitzGerald,  Dan Jurafsky, Hao Peng, Minjoon Seo, Noah A.~Smith, Swabha Swayamdipta, Sam Thomson, Trang Tran, Vicki Zayats and Luke Zettlemoyer for their helpful comments.

\bibliography{review}
\bibliographystyle{acl_natbib}

\renewcommand*\appendixpagename{\Large Appendices}

\clearpage
\begin{appendices}

\section{Acceptance Classification Features}\label{ssec:features}

Table \ref{tab:features} shows the features used by our acceptance classification model.
Figure \ref{fig:coef} shows the coefficients of all our features as learned by our best classifier on both datasets.

\subsection{Hyperparameters}\label{ssec:hyperparameters}
This section describes the hyperparameters used in our acceptance classification experiment.
Unless stated otherwise, we used the sklearn default hyperparameters. 
For decision tree and random forest, we used maximum depth=5.
For the latter, we also used max\_features=1.
For MLP, we used $\alpha=1$. 
For $k$-nearest neighbors, we used $k=3$.
For logistic regression, we considered both $l1$ and $l2$ penalty.

\begin{table*}[th]
\centering\small
\begin{tabular}{c|r|c|c}
\toprule
&\textbf{Features} & \textbf{Description} & \textbf{Labels} \\
\midrule
\parbox[t]{2mm}{\multirow{16}{*}{\rotatebox[origin=c]{90}{\textit{coarse}}}} 
&abstract\_contains\_{X} & \makecell{Whether abstract contains keywords X \\$\subset$ deep, neural, embedding, outperform, \\outperform, novel, state\_of\_the\_art } & boolean\\\cline{2-4}
&title\_length & \makecell{Length of title}& integer\\\cline{2-4}
&num\_authors & \makecell{Number of authors}& integer\\\cline{2-4}
&most\_recent\_refs\_year & \makecell{Most recent reference year} & 2001-2017\\\cline{2-4}
&num\_refs & \makecell{Number of references (\textit{sp})} & integer\\\cline{2-4}
&num\_refmentions & \makecell{Number of reference mentioned (\textit{sp})} & integer\\\cline{2-4}
&avg\_length\_refs\_mention & \makecell{Average length of references mentioned (\textit{sp})} & float\\\cline{2-4}
&num\_recent\_refs & \makecell{Number of recent references \\since the paper submitted (\textit{sp})}& integer\\\cline{2-4}
&num\_ref\_to\_X & \makecell{Number of X $\subset$ figures, tables, \\sections, equations, theorems (\textit{sp}) }  & integer  \\\cline{2-4}

&num\_uniq\_words & \makecell{Number of unique words (\textit{sp})}& integer\\\cline{2-4}
&num\_sections & \makecell{Number of sections (\textit{sp})}& integer\\\cline{2-4}
&avg\_sentence\_length & \makecell{Average sentence length (\textit{sp})}& float\\\cline{2-4}
&contains\_appendix & \makecell{Whether contains an appendix or not (\textit{sp})}& boolean\\\cline{2-4}
&prop\_of\_freq\_words & \makecell{Proportion of frequent words (\textit{sp})}& float\\\cline{2-4}
\hline
\parbox[t]{2mm}{\multirow{4}{*}{\rotatebox[origin=c]{90}{\textit{Lexical}}}}
&BOW & Bag-of-words in abstract& integer\\\cline{2-4}
&BOW+TFIDF & TFIDF weighted BOW in abstract& float\\\cline{2-4}
&GloVe & Average of GloVe word embeddings in abstract& float\\\cline{2-4}
&GloVe+TFIDF & \makecell{TFIDF weighted average \\of word embeddings in abstract} &float\\\cline{2-4}
\bottomrule
\end{tabular}
\caption{\label{tab:features} List of \textit{coarse} and \textit{lexical} features used for acceptance classification task. \textit{sp} refers features extracted from science-parse. }
\end{table*}

\begin{figure}[th]	
\centering
{
\includegraphics[trim=2cm 0.6cm 0cm 1cm,clip,width=.99\linewidth]{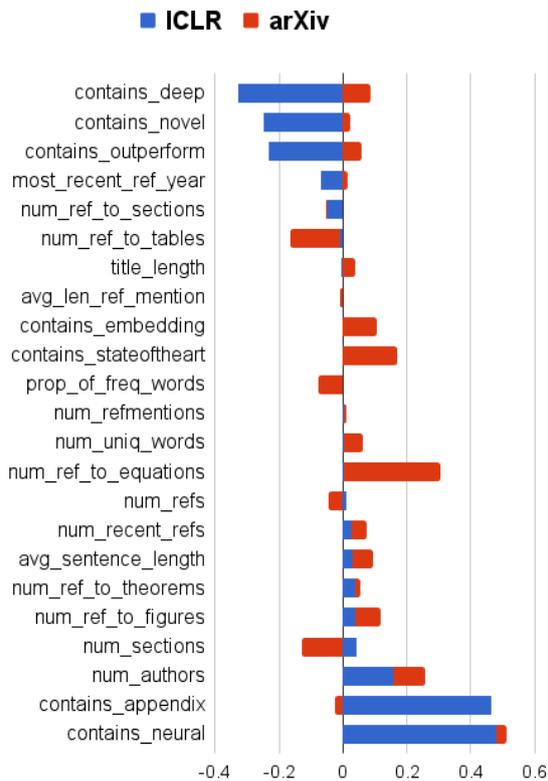}
}
\caption{\label{fig:coef} Coefficient values for coarse features in the paper acceptance classification, 
for {\color{blue}{ICLR}} and {\color{red}{arXiv}}.}
\end{figure}

\newpage
\newpage
\section{Reviewer Instructions}\label{ssec:instructions}
Below is the list of instructions to ACL 2016 reviewers on how to assign aspect scores to reviewed papers.

\begin{figure*}[th]
 \centering 
\includegraphics[scale=0.8]{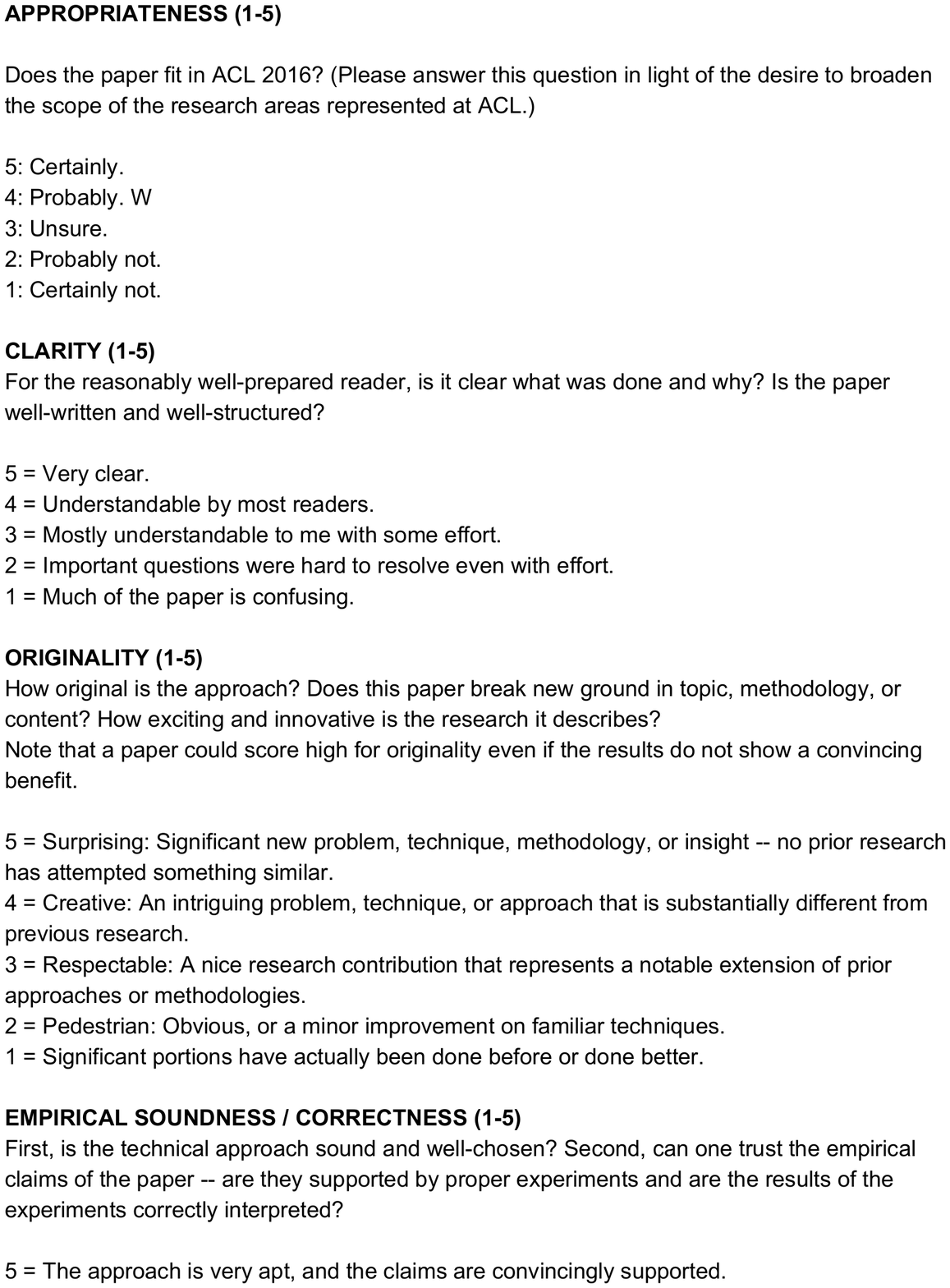}
\end{figure*}
\begin{figure*}[th]
 \centering 
\includegraphics[scale=0.8]{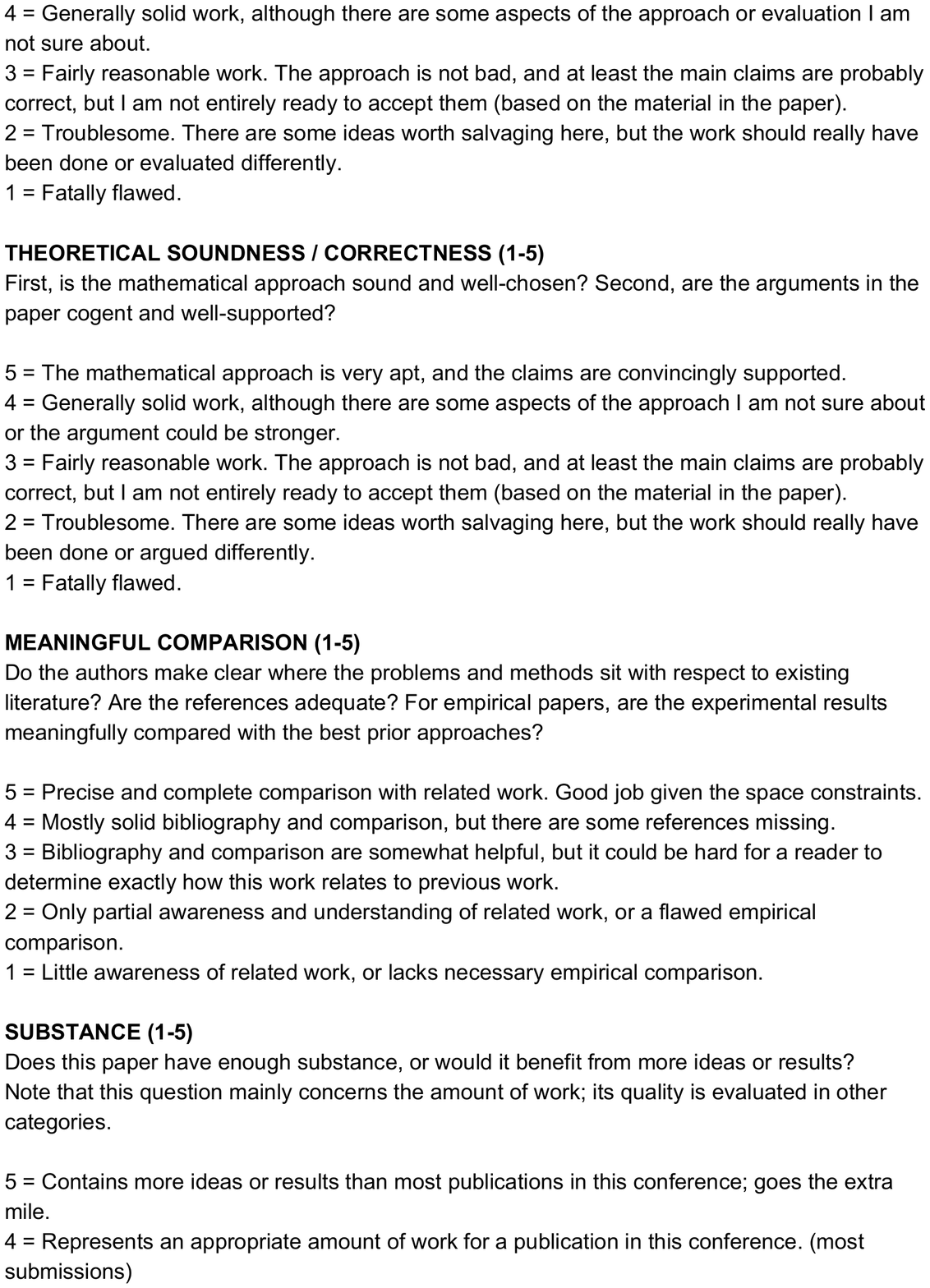}
\end{figure*}
\begin{figure*}[th]
 \centering 
\includegraphics[scale=0.8]{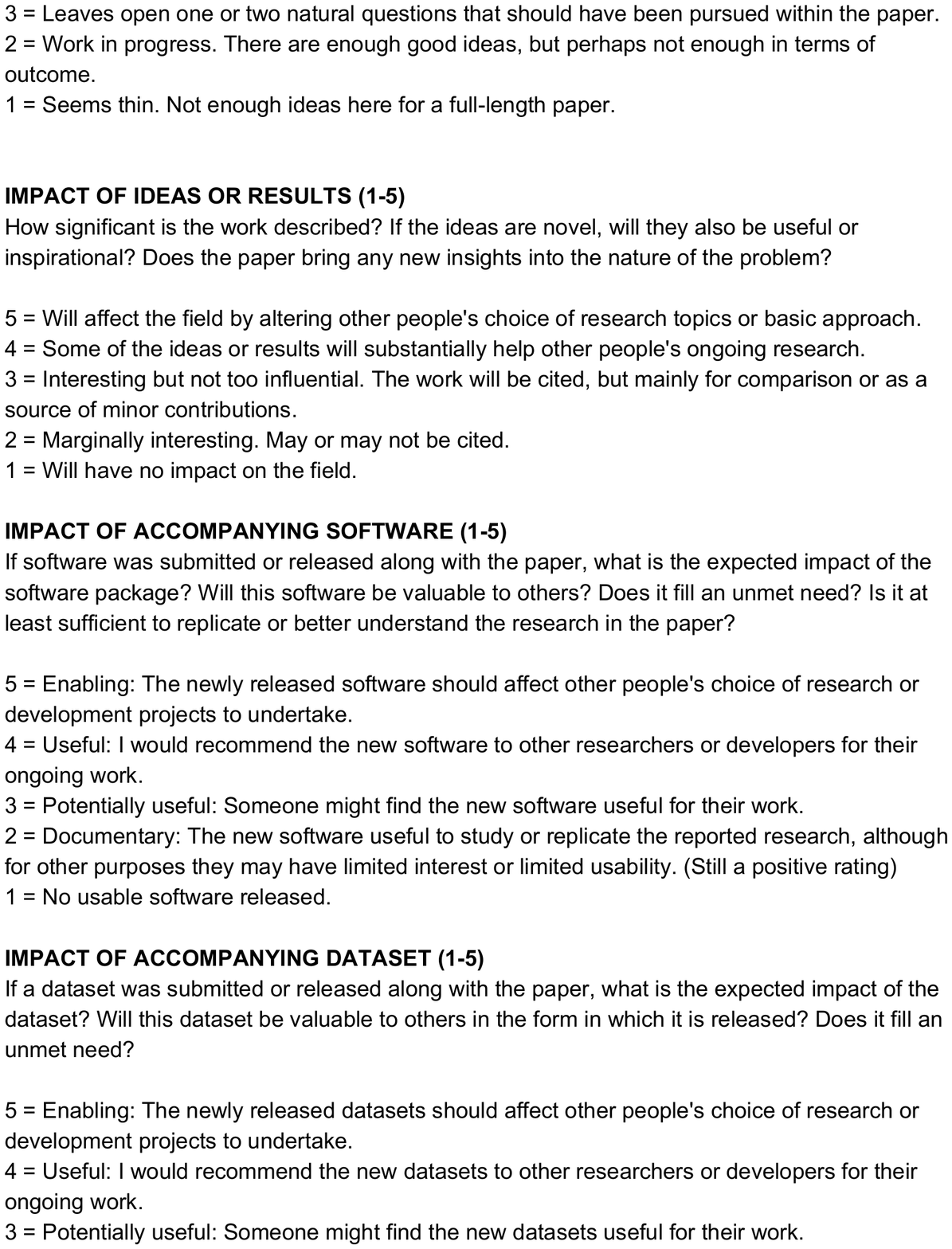}
\end{figure*}
\begin{figure*}[th]
 \centering 
\includegraphics[scale=0.8]{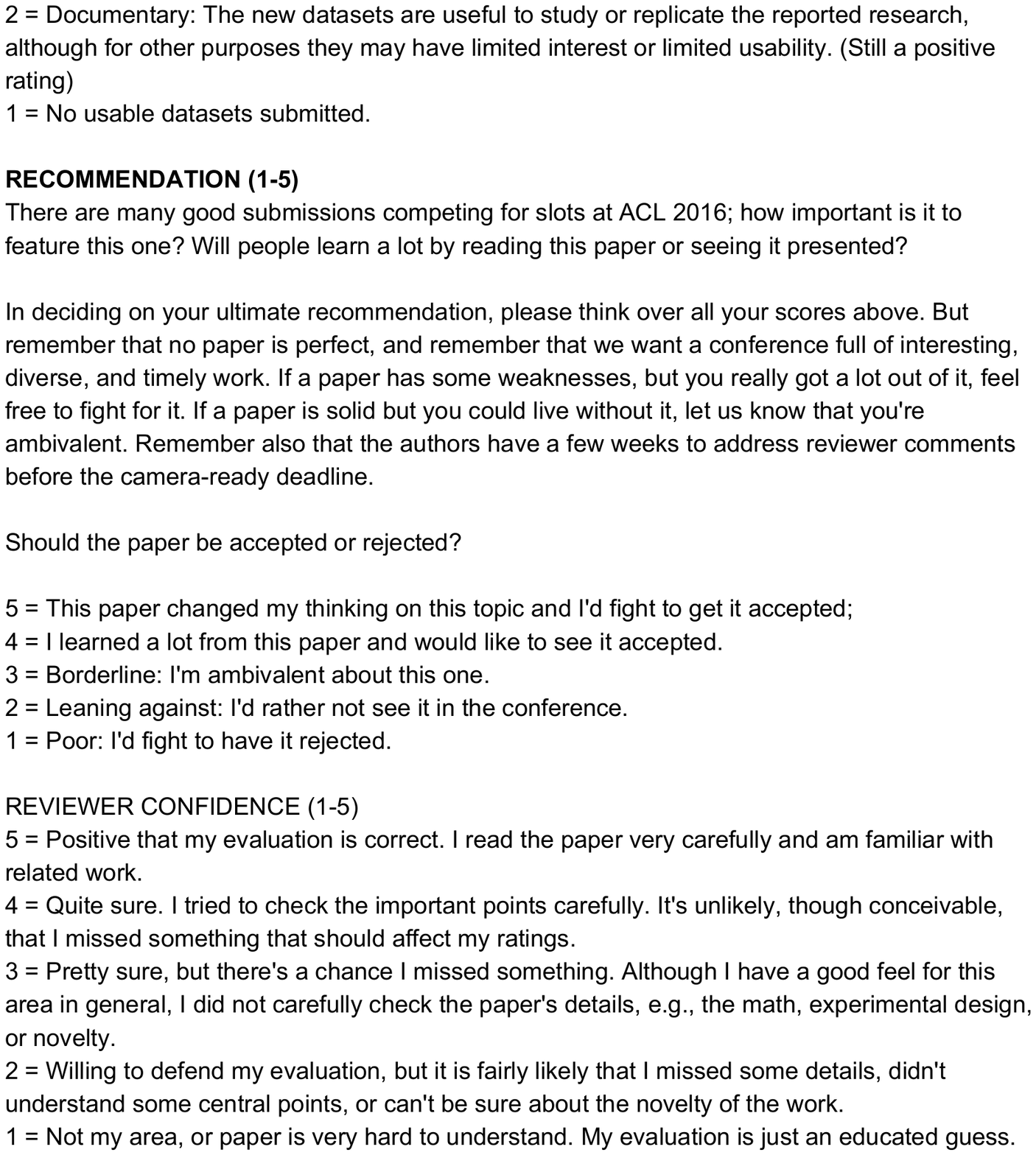}
\end{figure*}
\newpage
\com{
\begin{figure*}[th]	
\centering
{
\includegraphics[trim=5.9cm 0cm 6cm 0cm,clip,height=4.2cm,width=.99\linewidth]{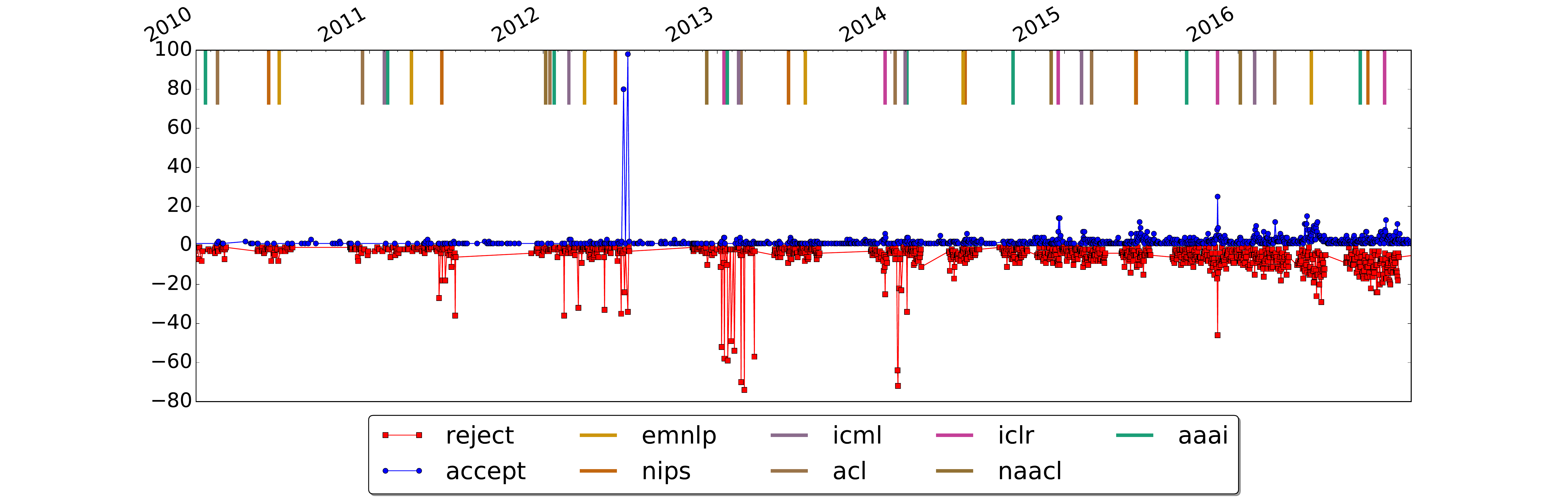}
}\
\caption{\label{fig:dist} Distribution of accepted (blue) / rejected (red) arXiv papers and submission dates of our target venues.}
\end{figure*}

\begin{figure}[h]	
\centering
{
\includegraphics[width=.9\linewidth]{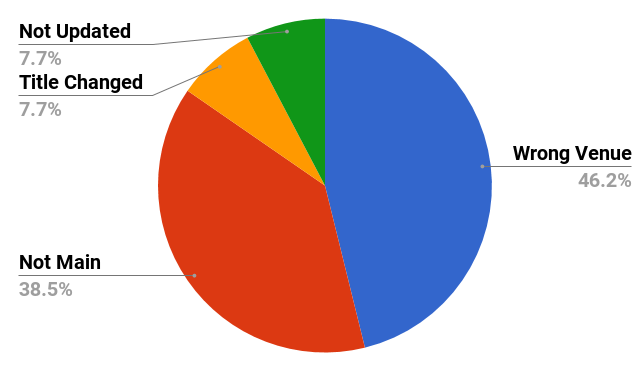}
}
\caption{\label{fig:sanity} Types of false negatives in our sanity check. [Wrong Venue] for 6, [Not Main] for 5, [Title Changed] for 1, and [Not Updated] for 1 paper.}
\end{figure}

Figure~\ref{fig:sanity} shows distribution of failure cases in our precision test.
The 6 papers in [Wrong Venue] and 5 papers [Not Main] are definitely negative samples.
The only one paper in [Title Changed] concatenate ``supplementary material'' to its title in the arXiv version so our matching algorithm didn't capture it. 
However, the supplementary document itself is a negative paper so it is not an issue.
One last paper in [Not Update] is detected but we can solve this problem by updating our crawling algorithm continuously.
Thus, basically we have only one false negative in [Not Update] case indeed.
You can find the full list of failure cases in the appendix section.

\com{
\begin{table*}
\footnotesize
\caption{\label{tab:failure} Full list of failure cases in sanity check. The comments are separated by ||| delimiter. }
\begin{tabularx}{\textwidth}{X}
\hline
\textbf{Title}: Character-Level Language Modeling with Hierarchical Recurrent Neural Networks\\
\textbf{Comments}: submitted to 29th conference on neural information processing systems (nips 2016) on may 20, 2016|||submitted to nips 2016 on may 20, 2016 (v1), accepted to icassp 2017 (v2) \\
\textbf{Failure}: [Different Venue] keywords matched but submitted to different venues \\
\hline
\textbf{Title}: How to scale distributed deep learning?\\
\textbf{Comments}: extended version of paper accepted at ml sys 2016 (at nips 2016)\\ 
\textbf{Failure}: [Not Main Conference] workshop paper\\
\hline
\textbf{Title}: Reinforcement Learning based Embodied Agents Modelling Human Users Through Interaction and Multi-Sensory Perception\\
\textbf{Comments}: 4 pages, 2 figures, accepted at the 2017 aaai spring symposium on interactive multi-sensory object perception for embodied agents|||4 pages, 2 figures, accepted at the 2017 aaai spring symposium on interactive multi-sensory object perception for embodied agents|||4 pages, 2 figures, accepted at the 2017 aaai spring symposium on interactive multi-sensory object perception for embodied agents\\ 
\textbf{Failure}: [Not Main Conference] workshop paper\\
\hline
\textbf{Title}: Getting deep recommenders fit: Bloom embeddings for sparse binary input/output networks\\
\textbf{Comments}: accepted for publication at acm recsys 2017; previous version submitted to iclr 2016\\ 
\textbf{Failure}: [Different Venue] keywords matched but submitted to different venues\\
\hline
\textbf{Title}: Parsing Natural Language Sentences by Semi-supervised Methods\\
\textbf{Comments}: dissertation interim report. overlap with papers accepted to acl 2015 and depling 2015, and a paper under review at iwpt 2015\\ 
\textbf{Failure}: [Different Venue] keywords matched but submitted to different venues\\
\hline
\textbf{Title}: Analysing Errors of Open Information Extraction Systems	\\
\textbf{Comments}: accepted at building linguistically generalizable nlp systems at emnlp 2017\\ 
\textbf{Failure}: [Not Main Conference] workshop paper\\
\hline
\textbf{Title}: Automatic Synthesis of Geometry Problems for an Intelligent Tutoring System\\
\textbf{Comments}: a formal version of the accepted aaai '14 paper\\ 
\textbf{Failure}: [Different Venue] keywords matched but submitted to different venues\\
\hline
\textbf{Title}: Cross-linguistic differences and similarities in image descriptions\\
\textbf{Comments}: accepted as a long paper for inlg 2017, santiago de compostela, spain, 4-7 september, 2017|||accepted for inlg 2017, santiago de compostela, spain, 4-7 september, 2017. camera-ready version. see the acl anthology for full bibliographic information\\ 
\textbf{Failure}: [Different Venue] keywords matched but submitted to different venues\\
\hline
\textbf{Title}: Transfer Learning for Neural Semantic Parsing\\
\textbf{Comments}: accepted for acl repl4nlp 2017\\ 
\textbf{Failure}: [Not Main Conference] workshop paper\\
\hline
\textbf{Title}: Multi-objective Reinforcement Learning with Continuous Pareto Frontier Approximation Supplementary Material\\
\textbf{Comments}: aaai-15 supplement. updated upon acceptance at the twenty-ninth aaai conference on artificial intelligence (aaai-15)\\ 
\textbf{Failure}: [Title Change] Multi-objective Reinforcement Learning with Continuous Pareto Frontier Approximation\\
\hline

\textbf{Title}: Understanding and Detecting Supporting Arguments of Diverse Types\\
\textbf{Comments}: this paper is accepted as a short paper in acl 2017\\ 
\textbf{Failure}: [Not Main Conference] short paper\\
\hline

\textbf{Title}: Learning Structured Text Representations\\
\textbf{Comments}: comments: 10 pages; typos corrected|||accepted by tacl\\ 
\textbf{Failure}: [Not updated] not updated to TACL proceedings\\
\hline

\textbf{Title}: Decision Trees for Function Evaluation - Simultaneous Optimization of Worst and Expected Cost\\
\textbf{Comments}: a preliminary version of this paper was accepted for presentation at icml 2014\\ 
\textbf{Failure}: [Different Venue] keywords matched but submitted to different venues\\
\hline
\end{tabularx}
\end{table*}

\begin{table}[h]
\centering\footnotesize
\caption{\label{tab:acceptance2} Root mean square error (RMSE) between encoders and between encoding text: Original scores and binarized scores on ACL and ICLR }
\begin{tabular}{@{}r@{}|c@{}c@{}c||c@{}c@{}c}
\toprule
& \multicolumn{3}{c}{Score}  & \multicolumn{3}{c}{Binary }  \\\hline
mean & \multicolumn{3}{c}{0.697}&\multicolumn{3}{c}{79.25}\\
Majority & \multicolumn{3}{c}{0.780}&\multicolumn{3}{c}{79.25}\\
\midrule
& $P(a|p)$ & $P(a|r)$ & $P(a|p,r)$ & $P(a|p)$ & $P(a|p,r)$ & $P(a|p,r)$ \\
\midrule
CNN & 0.767 & 0.725 & 0.795 & 74.07 & 77.03 & \textbf{80.0}\\
LSTM & 0.779 & 0.736 & 0.754 & 79.26 & 77.77 & 75.56\\
DAN & 0.740 & 0.751 & \textbf{0.723} & 77.04 & \textbf{80.0} & 77.78 \\
\bottomrule
\end{tabular}
\end{table}
}

}

\end{appendices}


\end{document}